\documentclass[10pt,twocolumn,letterpaper]{article}

\usepackage{cvpr}              

\usepackage{graphicx}
\usepackage{amsmath}
\usepackage{amssymb}
\usepackage{booktabs}
\usepackage[table]{xcolor}
\usepackage{adjustbox}

\usepackage[pagebackref,breaklinks,colorlinks]{hyperref}

\usepackage[capitalize]{cleveref}
\crefname{section}{Sec.}{Secs.}
\Crefname{section}{Section}{Sections}
\Crefname{table}{Table}{Tables}
\crefname{table}{Tab.}{Tabs.}

\def\cvprPaperID{*****} 
\def\confName{CVPR}
\def\confYear{2024}

\begin{document}

\title{Creative Portraiture: Exploring Creative Adversarial Networks and Conditional Creative Adversarial Networks}

\author{Sebastian Hereu \\
{\tt\small sebastian.hereu@columbia.edu}
\and
Qianfei Hu \\
{\tt\small qh2277@columbia.edu}
}
\maketitle

\begin{abstract}
Convolutional neural networks (CNNs) have been combined with generative adversarial networks (GANs) to create deep convolutional generative adversarial networks (DCGANs) with great success. DCGANs have been used for generating images and videos from creative domains such as fashion design and painting. A common critique of the use of DCGANs in creative applications is that they are limited in their ability to generate creative products because the generator simply learns to copy the training distribution. We explore an extension of DCGANs, creative adversarial networks (CANs). Using CANs, we generate novel, creative portraits, using the WikiArt dataset to train the network. Moreover, we introduce our extension of CANs, conditional creative adversarial networks (CCANs), and demonstrate their potential to generate creative portraits conditioned on a style label. We argue that generating products that are conditioned, or inspired, on a style label closely emulates real creative processes in which humans produce imaginative work that is still rooted in previous styles.
\end{abstract}

\section{Introduction/Motivation \& Related Work}
\label{sec:intro}

Society has long been fascinated with the idea of machine-based creativity. The study of machine-based creativity captures the public's imagination and inspires impassioned debate; some believe that creativity is a uniquely human quality, while others see a future where machines assist in and even automate creative processes. We believe research in the area of machine creativity has not only practical applications, but also deep philosophical implications.

GANs, introduced in [3], have been proven to be a powerful framework for estimating generative models. The unsupervised nature of GANs has great appeal in the field of Computer Vision, where labeled data is scarce and expensive to produce. 

Leveraging the unsupervised GAN architecture for Computer Vision tasks, the authors of [5] introduce DCGANs. The authors of [5] addressed the notoriously unstable training process of GANs by introducing architectural guidelines for stable DCGAN training, namely using strided convolutions instead of max pooling, batchnorm in generator and discriminator, and the LeakyReLU activation. DCGANs following this guideline have shown great promise for generative applications in Computer Vision.

Given the proper training data, DCGANs can generate products from creative domains such as fashion design and painting. Despite yielding impressive results, it may be said that these networks lack creativity as they simply learn to produce output resembling that of the training distribution. 

While humans themselves generate products by drawing on prior exposure and experience, they do so in a nuanced way that yields creative work. For AI to generate creative art, a quantifiable distinction -- i.e. one that can yield a loss function in the context of NNs -- must be made between creative synthesis and crude emulation. 

Towards this end, the authors of [2] drew on Colin Martingale's creativity hypothesis. According to Martingale, creative artists try to increase the arousal potential of their art to push against habituation. Yet, this arousal must be limited to avoid negative reactions by observers, who will have an aversion to art that departs too much from established styles and conventions.

The authors of [2] model this tension between convention and arousal in DCGANs by modifying the discriminator to have two heads: the traditional GAN head that decides whether the sample is real or fake, and a new head for placing a style label on the sample. The generator's loss is modified to consist of the same real/fake loss as before in addition to a cross-entropy loss that penalizes the generator for creating art that is easy to categorize as a specific style. To summarize, the idea is to generate art that is recognized as real by the discriminator but also hard to categorize as one specific style. According to the authors of [2], this new architecture and loss will yield more creative output than an unmodified DCGAN. 

In this paper, we explore the proposed CAN network, trained on portrait paintings from the WikiArt dataset, and compare it to a DCGAN baseline. Due to both experimental curiosity and hardware/time limitations, we train on scaled-down RGB images of size 64x64 instead of size 256x256 used in [2]. We show that our significantly pared-down network still achieves respectable results and demonstrates the merits of the CAN architecture. 

While the CAN architecture gives an interesting approach to generating creative products, we argue that it does not offer enough control over the generation process nor reflect the fact that artists are often inspired by a particular style when producing their creative works. For example, an artist may create an imaginative art piece inspired by the Impressionist art movement but is different enough to not belong to that category. To allow for the generation of creative products conditioned on a specific style class, we introduce the CCAN, which combines the ideas of [2] with the conditional GAN architecture of [4]. 

Like what was done for the DCGAN and CAN, we show selected output from our CCAN model. While the results of our CCAN are not as crisp as those of the baseline and CAN, we believe that given more hardware resources and longer training time, our CCAN model would be an interesting approach to generate creative, "inspired" artwork.
\section{Methodology}

As stated above, we wish to explore and evaluate the CAN and CCAN architectures for generating creative artwork. As a baseline for comparison, we will train an unmodified DCGAN. 

First, we compare the training stability of the models via their loss graphs. This is important as training stability is a major concern in GANs and may be said to be a primary limiting factor for GANs. Demonstrating the training stability of CAN and CCAN models is important if these models are to be used in practice. 

Next, we present and qualitatively evaluate the output of the models. Of course, creativity is entirely subjective. What we can evaluate, however, is whether the CAN and CCAN produce artwork that is discernibly different from the baseline GCAN. To demonstrate the CANs creativity, the authors of [2] performed a blind study in which participants evaluated the output of a baseline, the CAN, and real artwork for creativity. We do not have access to such resources, so we are constrained to qualitative evaluations. Nevertheless, since human participants were used in the original study to evaluate the creativity of the artwork, our discussion of the qualitative results of the artwork as well as the readers' evaluation will serve as a reasonable proxy. 

Like the authors of [2], we use the WikiArt dataset. The WikiArt dataset is a large dataset containing 81,444 pieces of visual art from various artists. The images of art are taken from WikiArt.org. Each image is labeled with a "style" tag, which specifies which style that particular piece belongs to. See Table 1 for a list of the styles in the training dataset.

While [2] uses the entire WikiArt dataset, we limit the training data to only portrait images. In addition to speeding up training time, using artwork with a clear (human) figure limits the subject matter of the artwork as a confounding variable for evaluating the creativity of the output. Otherwise, as can be seen in [2], much of the artwork is abstract, and it may be difficult to differentiate between a cohesive artwork and garbled, randomized colors. We hope that forcing the CAN and CCAN to generate artwork with a central human figure will lead the models to find other means of introducing creativity into the artwork, such as through expression, pose, and clothing. After selecting all portrait images, our training dataset contained 14,245 data points. Several images from our training dataset are shown below.
\includegraphics[width=8cm, height=8cm]{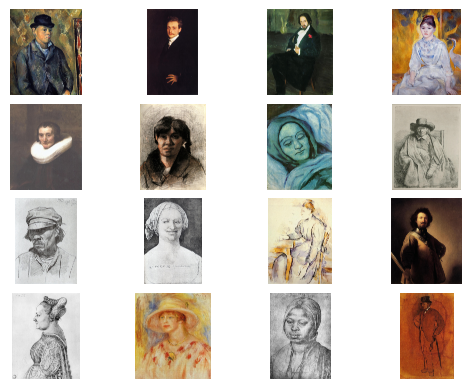}
  
We took the initial preprocessing step of normalizing the images so that their values were between -1 and 1. This is a necessary step to work with the tanh activation in the output layer of the generator. Next, we performed a five crop for each image. That is, for each of the 14,245 portraits, we created five crops of size $(0.9*h, 0.9*w)$, where $h$ and $w$ are the original height and width of the image. The five crops generated for each image include a central crop taken from the center of the image and four corner crops taken from each corner of the image. We believe this cropping step is essential, especially for our reduced-size model, as otherwise, the model may have learned idiosyncrasies of the training data instead of the salient patterns of the artwork.

In [2], the authors resized the training images to 256x256 to use as input into their models. Initially, we used this size as well, but later found that the required models were taking too long to train, considering the resource and time constraints for this project. For example, using Google Colab and an A100 GPU with 40GB of RAM, 30 hours of training was not enough to generate images that even remotely resembled the shape of a human figure. Thus, we decided to scale down the images to 64x64. Doing so allowed us to train, debug, and iterate the models quickly and effectively. We believe that our 64x64 results still demonstrate the merits of the CAN and CCAN architectures, and future researchers may scale up the network further for more detailed results.

\begin{table}[htbp]
    \centering
    \caption{Art style tags in WikiArt}
    \small 
    \setlength{\tabcolsep}{10pt} 
    \begin{adjustbox}{max width=\textwidth} 
        \begin{tabular}{|*{2}{c|}}
            \hline
            \rowcolor{gray!30}
            \textbf{} & \textbf{} \\
            \hline
            Abstract\_Expressionism & Action\_painting \\
            Analytical\_Cubism & Art\_Nouveau \\
            Baroque & Color\_Field\_Painting \\
            Contemporary\_Realism & Cubism \\
            Early\_Renaissance & Expressionism \\
            Fauvism & High\_Renaissance \\
            Impressionism & Mannerism\_Late\_Renaissance \\
            Minimalism & Naive\_Art\_Primitivism \\
            New\_Realism & Northern\_Renaissance \\
            Pointillism & Pop\_Art \\
            Post\_Impressionism & Realism \\
            Rococo & Romanticism \\
            \hline
        \end{tabular}
    \end{adjustbox}
    \label{tab:wikiart_tags}
\end{table}

Next, we describe the models themselves. We will describe the loss functions and training afterward. Our DCGAN baseline, although significantly reduced in complexity compared to that used in [2], follows the prescription for stable DCGANs described in [5]. See Table 2 for the architecture. 

\begin{table}[htbp]
    \centering
    \caption{DCGAN architecture}
    \footnotesize
    \resizebox{0.6\columnwidth}{!}{%
    \begin{tabular}{lll}
        \toprule
        \multicolumn{3}{c}{\textbf{Generator}} \\
        \midrule
        Layer (type) & Output Shape & Param \# \\
        \midrule
        ConvTranspose2d-1 & [-1, 512, 4, 4] & 819,200 \\
        BatchNorm2d-2 & [-1, 512, 4, 4] & 1,024 \\
        ReLU-3 & [-1, 512, 4, 4] & 0 \\
        ConvTranspose2d-4 & [-1, 256, 8, 8] & 2,097,152 \\
        BatchNorm2d-5 & [-1, 256, 8, 8] & 512 \\
        ReLU-6 & [-1, 256, 8, 8] & 0 \\
        ConvTranspose2d-7 & [-1, 128, 16, 16] & 524,288 \\
        BatchNorm2d-8 & [-1, 128, 16, 16] & 256 \\
        ReLU-9 & [-1, 128, 16, 16] & 0 \\
        ConvTranspose2d-10 & [-1, 64, 32, 32] & 131,072 \\
        BatchNorm2d-11 & [-1, 64, 32, 32] & 128 \\
        ReLU-12 & [-1, 64, 32, 32] & 0 \\
        ConvTranspose2d-13 & [-1, 3, 64, 64] & 3,072 \\
        Tanh-14 & [-1, 3, 64, 64] & 0 \\
        \midrule
        \multicolumn{3}{c}{\textbf{Discriminator}} \\
        \midrule
        Conv2d-1 & [-1, 64, 32, 32] & 3,072 \\
        LeakyReLU-2 & [-1, 64, 32, 32] & 0 \\
        Conv2d-3 & [-1, 128, 16, 16] & 131,072 \\
        BatchNorm2d-4 & [-1, 128, 16, 16] & 256 \\
        LeakyReLU-5 & [-1, 128, 16, 16] & 0 \\
        Conv2d-6 & [-1, 256, 8, 8] & 524,288 \\
        BatchNorm2d-7 & [-1, 256, 8, 8] & 512 \\
        LeakyReLU-8 & [-1, 256, 8, 8] & 0 \\
        Conv2d-9 & [-1, 512, 4, 4] & 2,097,152 \\
        BatchNorm2d-10 & [-1, 512, 4, 4] & 1,024 \\
        LeakyReLU-11 & [-1, 512, 4, 4] & 0 \\
        Conv2d-12 & [-1, 1, 1, 1] & 8,192 \\
        Sigmoid-13 & [-1, 1, 1, 1] & 0 \\
        \midrule
        Total params: & \multicolumn{2}{l}{6,342,273} \\
        \bottomrule
    \end{tabular}%
    }
    \label{tab:model-architecture}
\end{table}
Note that we reshape the output of the discriminator to have size [-1, 1], reflecting the discriminator's predicted probability that a sample is real or fake. 

The only difference between the above DCGAN and the CAN is that the CAN contains an additional head in the discriminator that takes the output from Conv2d-9. See Table 3 for a description of the second head. Note the softmax activation in the output layer has shape [-1, 24], reflecting the predicted probability that a sample belongs to one of 24 style classes.

\begin{table}[htbp]
    \centering
    \caption{Multi-label Probabilities Head for CAN}
    \footnotesize
    \resizebox{0.6\columnwidth}{!}{%
    \begin{tabular}{lll}
        \toprule
        Layer (type) & Output Shape & Param \# \\
        \midrule
        Flatten-14 & [-1, 8192] & 0 \\
        Linear-15 & [-1, 1024] & 8,389,632 \\
        LeakyReLU-16 & [-1, 1024] & 0 \\
        Linear-17 & [-1, 512] & 524,800 \\
        LeakyReLU-18 & [-1, 512] & 0 \\
        Linear-19 & [-1, 24] & 12,312 \\
        Softmax-20 & [-1, 24] & 0 \\
        \bottomrule
    \end{tabular}%
    }
    \label{tab:style_class_head}
\end{table}

The CCAN architecture is essentially the CAN architecture with the addition of class-tag embeddings described in [4] that allow for class-conditional generation. We tried many methods of adding class-tag embeddings to the CAN model initially to no avail. Finally, we found a particular configuration that led to stable training and coherent output. In the generator, we create an embedding that maps a style label to a 100-dimensional vector and concatenates that vector with the latent noise vector. In the discriminator, we add an embedding that maps a style label to a 3-dimensional vector and concatenates the embedding with the input sample.

For the DCGAN, we used the traditional loss function for GANs described in [3]. The loss function (from [2]) we used to train the CAN reflects the goal of the generator: to create artwork that is recognized as real by the discriminator yet is not categorizable as a specific style:

\begin{small}
\begin{align*}
\min_{G} \max_{D} V(D, G) = 
& \mathbb{E}_{x,c \sim p_{\text{data}}} [\log D_r(x) + \log D_c(c = \hat{c}|x)] \\
& + \mathbb{E}_{z \sim p_z} [\log(1 - D_r(G(z))) - \sum_{k=1}^{K} (\frac{1}{K} \log D_c(c_k|G(z)) \\
& + \left(1 - \frac{1}{K}\right) \log(1 - D_c(c_k|G(z))))]
\end{align*}
\end{small}

Note that the subtracted term above is the fake-image cross-entropy loss, $z$ is a noise vector sampled from the distribution $p_z$, $x$ and $\hat{c}$ are the real image and its style label, $D_r(\cdot )$ is the function (head) that tries to discriminate between real and generated (fake) images, and $D_c(\cdot )$ is the function (head) that discriminates between different style categories and estimates the style class. See Algorithm 1 (Figure 1), which describes how the CAN is trained. Note that Algorithm 1 is directly taken from [2].

\begin{figure}[t]
    \centering
    \includegraphics[width=1.0\linewidth,height=2.3in]{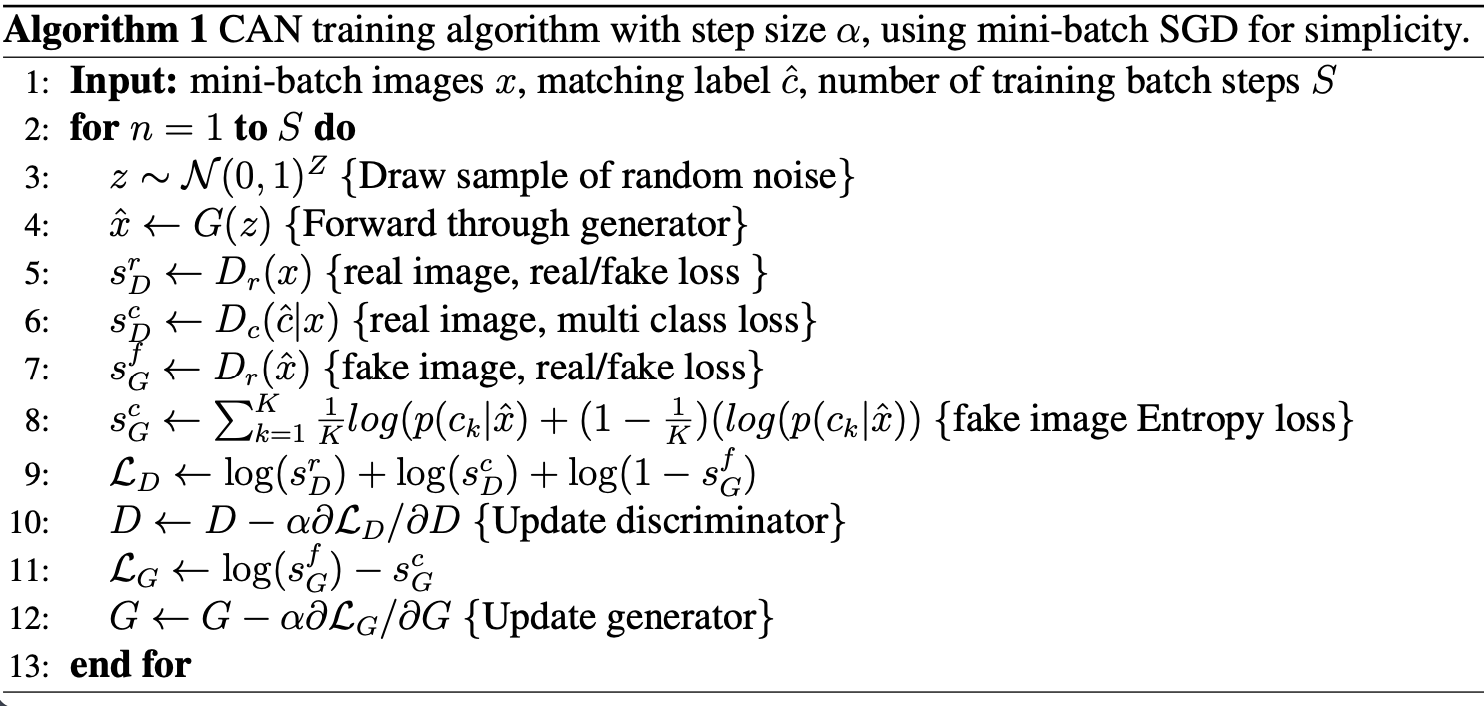}
    \caption{Algorithm 1. showing how the CAN is trained. Note that the CCAN is trained in a very similar manner, with the only addition being that style labels are provided to the discriminator and generator.}
\end{figure}

Note that the CCAN uses the same loss function, with the minor difference that the discriminator and generator are provided with style labels while training. Please refer to our implementation notebooks, provided with the submission, for more details about the training process.

For all models, we trained for 120 epochs, used a batch size of 128, and used the Adam optimizer with learning rate $0.0001$, $\beta_1 = 0.5$, $\beta_2 = 0.999$. The training process took 24hours for each model on an Nvidia V-100 GPU in the Google Colab environment. 

\section{Experimental Results \& Discussion}

Compared to the baseline DCGAN, the CAN and CCAN displayed similar stability. Looking at Figure 2, it can be seen that three models display similar loss-shapes. It should be noted that although the discriminator loss was decreasing while the generator loss was increasing over 120 epochs for all three models, this occurred slowly, and the discriminator loss remained well above zero. Thus, mode collapse was averted for all three models.

\begin{figure}[t]
    \centering
    \fbox{\includegraphics[width=1.07\linewidth,height=2.3in]{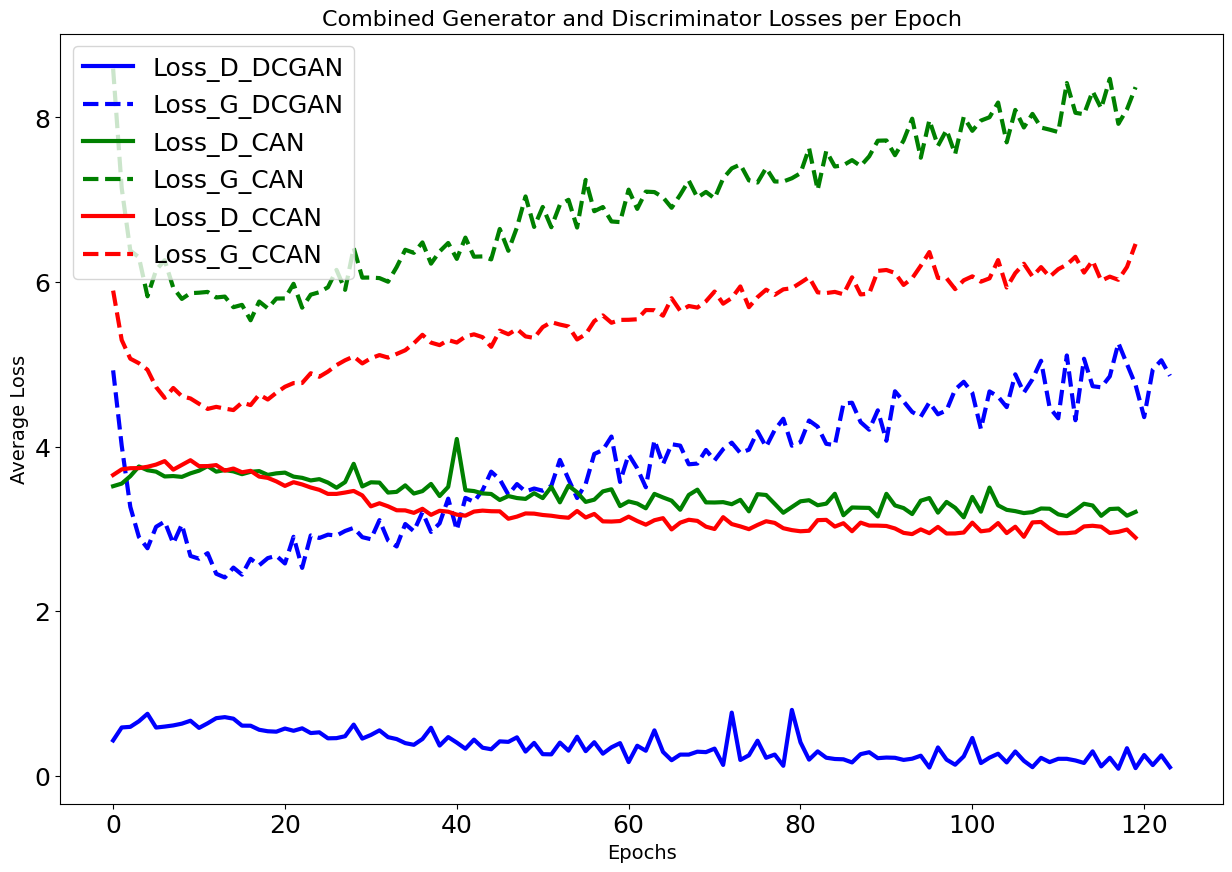}}
    \caption{Average Generator and Discriminator losses over 120 epochs. Training was stable for all three epochs and the Discriminator and Generator loss did not diverge over 120 epochs}
    \label{fig:onecol}
\end{figure}

In Figures 3, 4, and 5, we show selected outputs for the DCGAN, CAN, and CCAN models respectively. The DCGAN produced impressive portraits (Figure 3) that show a variety of poses, ranging from head shots to full-body shots. Many of the facial features of the figures appear crude and emotionless, but are nevertheless impressive considering the small size of the model. Most figures appear to have similar outfits and headdresses, resembling the Baroque-style examples in the training set. This makes sense given there is little incentive for the DCGAN to produce diverse output.  

In our opinion, the CAN's output (Figure 4) displays a broader and more interesting range of portraiture than the DCGAN. For example, consider the figure in the upper left-hand corner of the CAN output collage. The style of the figure does not resemble any of the other figures from the CGAN or CAN outputs, with its white headdress and hauntingly gaunt face. Moreover, we believe that the facial features in the CAN output tend to be more elaborate and emotive. For example, none of the figures produced by the DCGAN rarely, if ever, have facial hair, while the CAN often produces portraits of figures with facial hair. Consider the figure in the lower left-hand corner of the CAN output, who appears to have a mustache. 

The CCAN appears to have generated class-conditional, yet 'creative' images (Figure 5). Note that each row in the CCAN collage represents a different class style. Consider the second row. The portraits appear to be painted in a circular frame, reflecting many of the portraits of that style in training data. We note that the portraits appear to have fewer facial features than the DCGAN and CAN outputs, but argue that the style of portraits produced still captures the essential elements of the conditioned-on style label. For each style label, a variety of interesting pieces are generated, each with unique formal qualities, ranging from color, lighting, shading, and pose. We believe that a more complex CCAN architecture and longer training epochs could produce more detailed facial features.

That we successfully trained a model both conditioning on class labels and using fake-image entropy loss seems paradoxical: on one hand, conditioning on a class moves the generator towards generating portraits with that style label, on the other hand, the fake-image entropy loss makes the generator create portraits that do not fall in a specific style. We argue that this tension produced creative outputs, as the model was forced to find subtle and interesting ways to create style ambiguity while still being rooted in a specific style. Moreover, it may be said that this tension closely reflects the tension between creativity and arousal described by Martingale, and is a new interesting approach to creative art generation that builds upon [2].

\begin{figure}[t]
    \centering
    \fbox{\includegraphics[width=1\linewidth,height=3in]{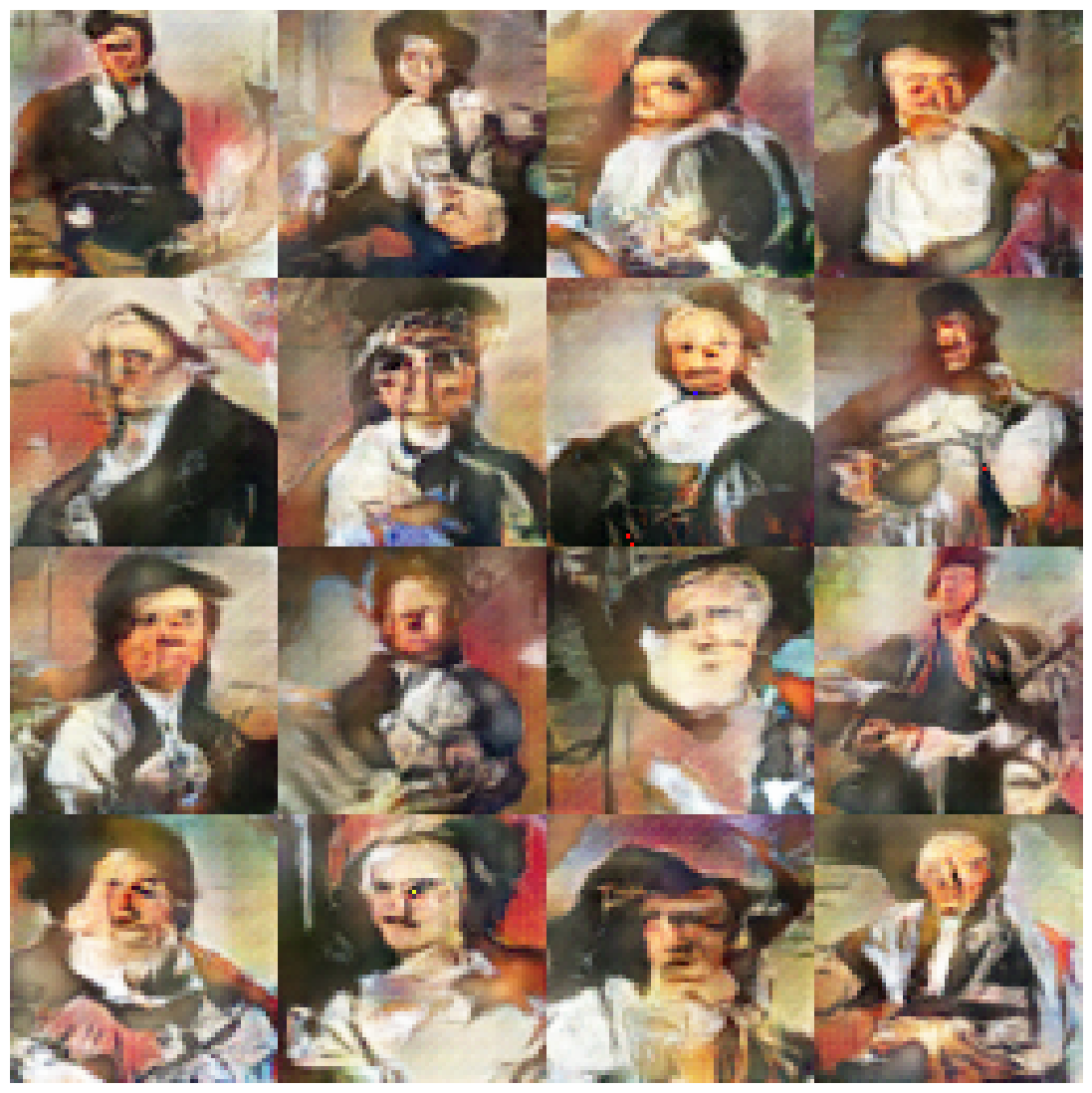}}
    \caption{Selected output from DCGAN. Note the impressive qualities of the output, including the positioning of the central figure, the clothing details. Notice the limited and crude facial expressions.}
    \label{fig:onecol}
\end{figure}

\begin{figure}[t]
    \centering
    \fbox{\includegraphics[width=1\linewidth,height=3in]{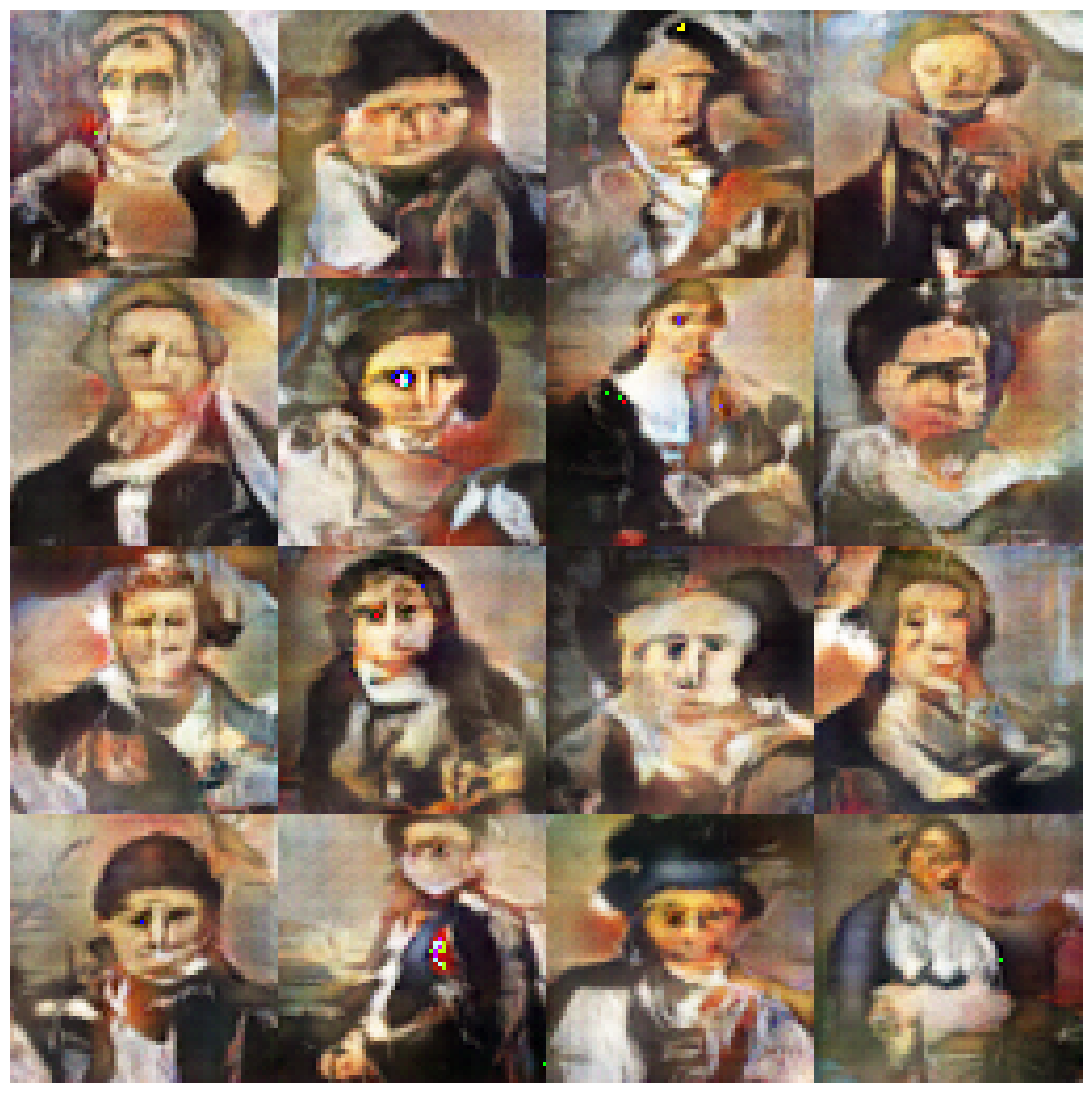}}
    \caption{Selected output from CAN. Note the diverse style of the output when compared to the DCGAN}
    \label{fig:onecol}
\end{figure}

\begin{figure}[t]
    \centering
    \fbox{\includegraphics[width=1\linewidth,height=3in]{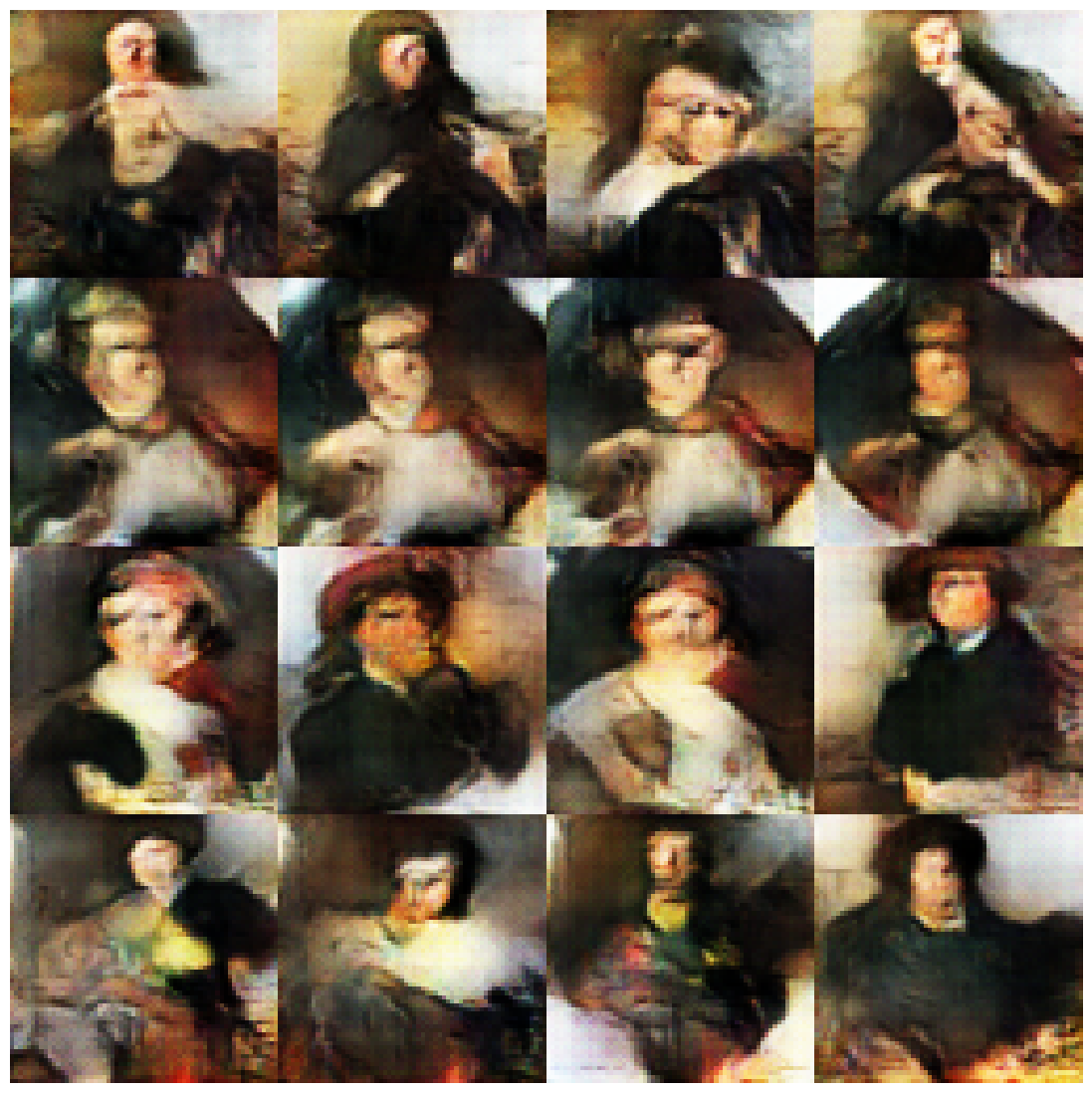}}
    \caption{Selected output from CCCAN. Each row is the resultant output of a different style vector.
    Row 1: Realism, Row 2: Rococo, Row 3: Romanticism, Row 4: Baroque. Note that the portraits in each row have similar styles, yet differing details}
    \label{fig:onecol}
\end{figure}

\section{Conclusion}

To summarize, we implemented, trained, and evaluated a DCGAN baseline, a CAN, and our proposed architecture CCAN on portraits from the WikiArt dataset. The DCGAN and CAN were significantly reduced in their complexity compared to [2] and still produced interesting, artistic portraits; the original model had over 50 million parameters while our implementations of CAN and DCGAN contained only about 6 million parameters. 

The CAN produced portraits that were divergent in appearance and style from those of the DCGAN, showing that the CAN architecture is a viable means of generating coherent yet style-ambiguous art. 

We also successfully trained and demonstrated the outputs of our proposed CCAN model, which combines the ideas from [2] and [4] to generate class-conditional style-ambiguous art. We noted that the outputs of that model did not have as much detail as the DCGAN and CAN outputs.

For future work, we believe that given a more complex network and larger output size, our CCAN could produce far more detailed output. In any case, our CCAN produced output of artistic merit and may serve as a proof-of-concept for future researchers who have the time and computational resources to implement a scaled-up CCAN. 

The open question remains if machines can truly be creative. The CAN and CCAN still generate art using essentially a mixture of the training data. Modeling human arousal and the novelty of AI-generated works presents a significant challenge, requiring a deeper understanding of human perception, emotion, and aesthetic preferences. Achieving creativity in machines may require advances in cognitive science and neuroscience to develop AI systems capable of true imaginative thinking.

\section{Code}

Github Repository: \url{https://github.com/sebastianhereu/cv2_proj/} \\ 

The above repository contains: \\

1) Notebooks that contain the definitions and training of the models. \\

2) Google Drive links to the Pytorch state dictionaries of the final trained models and the dataset \\

3) An example notebook demonstrating portrait generation using the pre-trained models. \\



{\small

\cite{1406.2661}, \cite{1411.1784}, \cite{1511.06434}, \cite{1706.07068}, \cite{wikiart_dataset}.
\bibliographystyle{ieee_fullname}
\bibliography{egbib}
}

\section{Contributions}

Implementing DCGAN \& Training: Ben, Sebastian \\

Implementing CAN \& Training: Ben, Sebastian \\

Implementing CCAN \& Training: Ben, Sebastian \\

Report: Ben, Sebastian \\

\end{document}